\documentclass[letterpaper, 10 pt, conference]{ieeeconf}  

\IEEEoverridecommandlockouts                              

\usepackage{siunitx}
\usepackage{ifthen}
\usepackage{booktabs}
\usepackage{multirow}
\usepackage{graphicx}
\usepackage{cite}
\usepackage{caption}
\usepackage{subcaption}
\usepackage{bm}
\usepackage{amsmath}
\usepackage{amsfonts}
\usepackage{float}
\usepackage{varioref}
\usepackage{paralist}
\usepackage{cleveref}
\crefformat{section}{\S#2#1#3} 
\crefformat{subsection}{\S#2#1#3}
\crefformat{subsubsection}{\S#2#1#3}

\newcommand{\thisWork}{R-ALC}
\newcommand{\alcv}{ALCv2}

\newcommand{\ShowComments}{true} 
\newcommand{\SR}[1]{
    \ifthenelse{\equal{\ShowComments}{true}}{\noindent\textcolor{blue}{Sadegh: #1}}{}
}

\title{\LARGE \bf
Region Based SLAM-Aware Exploration: Efficient and Robust Autonomous Mapping Strategy That Can Scale
}

\author{Megha Maheshwari$^\star$, Sadegh Rabiee$^\star$, He Yin$^\star$, Martin Labrie$^\star$, Hang Liu, Rajasimman Madhivanan
\thanks{All authors are with Amazon Lab126.} 
\thanks{\tt\scriptsize{\{mahemegh,srabiee,heyinz,labrieml,luhang\}@amazon.com}, rajasimm@lab126.com}
\thanks{$^\star$ authors contributed equally.}
}

\begin{document}

\maketitle
\thispagestyle{empty}
\pagestyle{empty}

\begin{abstract}

Autonomous exploration for mapping unknown large scale environments is a fundamental challenge in robotics, with efficiency in time, stability against map corruption and computational resources being crucial. This paper presents a novel approach to indoor exploration that addresses these key issues in existing methods. We introduce a Simultaneous Localization and Mapping (SLAM)-aware region-based exploration strategy that partitions the environment into discrete regions, allowing the robot to incrementally explore and stabilize each region before moving to the next one. This approach significantly reduces redundant exploration and improves overall efficiency. As the device finishes exploring a region and stabilizes it, we also perform SLAM keyframe marginalization, a technique which reduces problem complexity by eliminating variables, while preserving their essential information. To improves robustness and further enhance efficiency, we develop a checkpoint system that enables the robot to resume exploration from the last stable region in case of failures, eliminating the need for complete re-exploration. Our method, tested in real homes, office and simulations, outperforms state-of-the-art approaches. The improvements demonstrate substantial enhancements in various real world environments, with significant reductions in keyframe usage (85\%), submap usage (50\% office, 32\% home), pose graph optimization time (78-80\%), and exploration duration (10-15\%). This region-based strategy with keyframe marginalization offers an efficient solution for autonomous robotic mapping.

\end{abstract}

\section{INTRODUCTION}

Autonomous exploration and mapping of unknown environments are prerequisites for various mobile robot applications, including search and rescue operations and service robotics. While significant progress has been made in autonomous mobile exploration \cite{bircher2016receding, dang2019graph}, efficient exploration of large-scale environments remains a complex task due to the following reason.
\begin{enumerate}
    \item 
    Time taken to explore the entire floor plan: One of the limitations of frontier based exploration is the tendency for the robot to revisits the same area multiple times to close unexplored frontiers which is inefficient as it leads to increased exploration time. 
    \item 
    Map drift during prolonged exploration: As the device continues to explore, map drift can occur, becoming more significant as the mapped area expands. This drift makes periodic map stabilization essential. While active loop closure helps by periodically connecting previously visited locations, it doesn't guarantee complete stabilization of the entire explored map.
     \item 
     Increased Memory Usage: As the device explores the number of generated keyframes, as well as the number of submaps increase which lead to increased memory usage. Furthermore the map size also grows leading to increased memory.
     \item 
     Recovery from failures: With the current approaches, e.g. \cite{ALCv2, Mohit2023lighthouse}, in case of failures, there is no efficient way to resume exploration from the point of failure. When the device is performing exploration and it fails for any reason, the current way is to resume exploration from the start. One straightforward way would be to save the map that is already explored and resume from there. However, stability and quality of this saved map is not guaranteed. Hence the point from which exploration should resume after failure continues to be a challenge.
\end{enumerate}
\begin{figure}[t]
  \centering
  \includegraphics[width=0.45\textwidth]{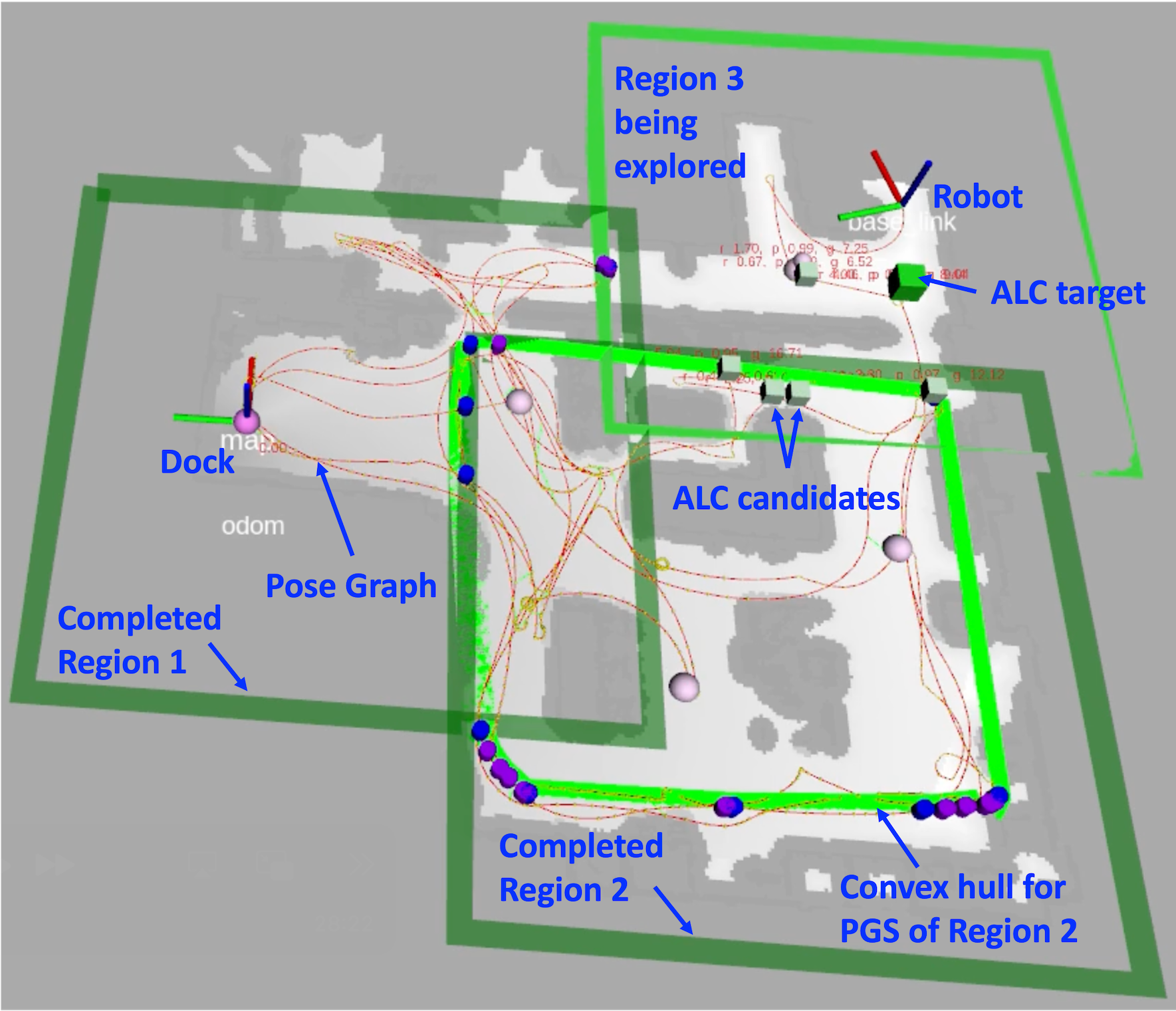}
  \caption{\textbf{Region based SLAM-aware (\thisWork{}) Exploration}. Our strategy explores the environment region by region. Planners ensure each region is fully explored and stabilized before moving to the next, preventing unnecessary back-and-forth movement during exploration, and improving the efficiency. Light green rectangles indicate regions under exploration, while dark green ones show completed regions. Once a region is completed, the SLAM system freezes its poses for future optimization, preventing relative motions within this region while allowing global transformations. This approach significantly reduces computational and memory requirements. In the case of exploration failure, the system loads the latest completed region, enabling resumption from a stable state.}
\end{figure}

In order to overcome these challenges we present a novel region based SLAM-aware exploration algorithm (\thisWork{}) that addresses the above challenges using the following approach and in addition also proposes an overall system design. The overall contribution of the paper are as follows:
\begin{enumerate}
    \item 
    Region Based Exploration: In order to make exploration more efficient in terms of time and memory our approach partitions the environment into discrete regions and each region is explored and stabilized before exploring another. After all regions are explored, a global stabilization is performed to stabilize the entire map.
    \item 
    Pose Graph Marginalization: Once a region is explored and stabilized we marginalize a subset of the keyframes within it. This helps in reduction of compute and memory as we do not need to re-process the information from the stabilized region.
    \item 
    Exploration Recovery: To resume exploration from a good quality map, our approach stabilizes each region after it is fully explored. This allows the flexibility to resume exploration from the last stabilized region in case of failure.
\end{enumerate}
We demonstrate the effectiveness of our proposed strategy through comprehensive on-device testing and simulation experiments. The results show significant improvements in exploration efficiency, while being resilient to errors compared to traditional exploration methods. 


\section{Related Work}
\begin{figure*}
     \centering
     \begin{subfigure}[b]{0.22\textwidth}
         \centering
         \includegraphics[width=\textwidth]{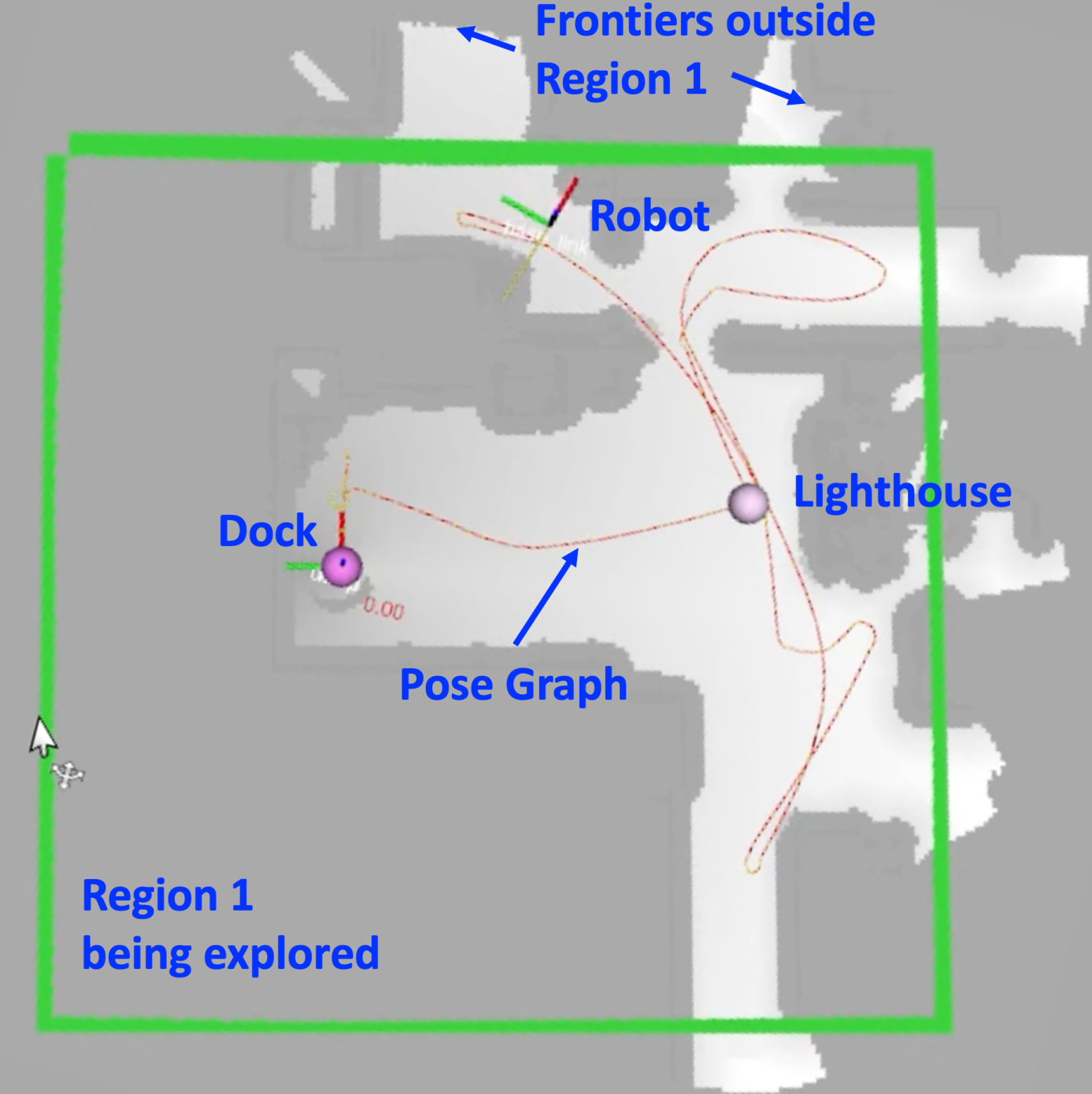}
         \caption{Region discovery of Region 1}
         \label{fig:step1}
     \end{subfigure}
     \hfill
     \begin{subfigure}[b]{0.22\textwidth}
         \centering
         \includegraphics[width=\textwidth]{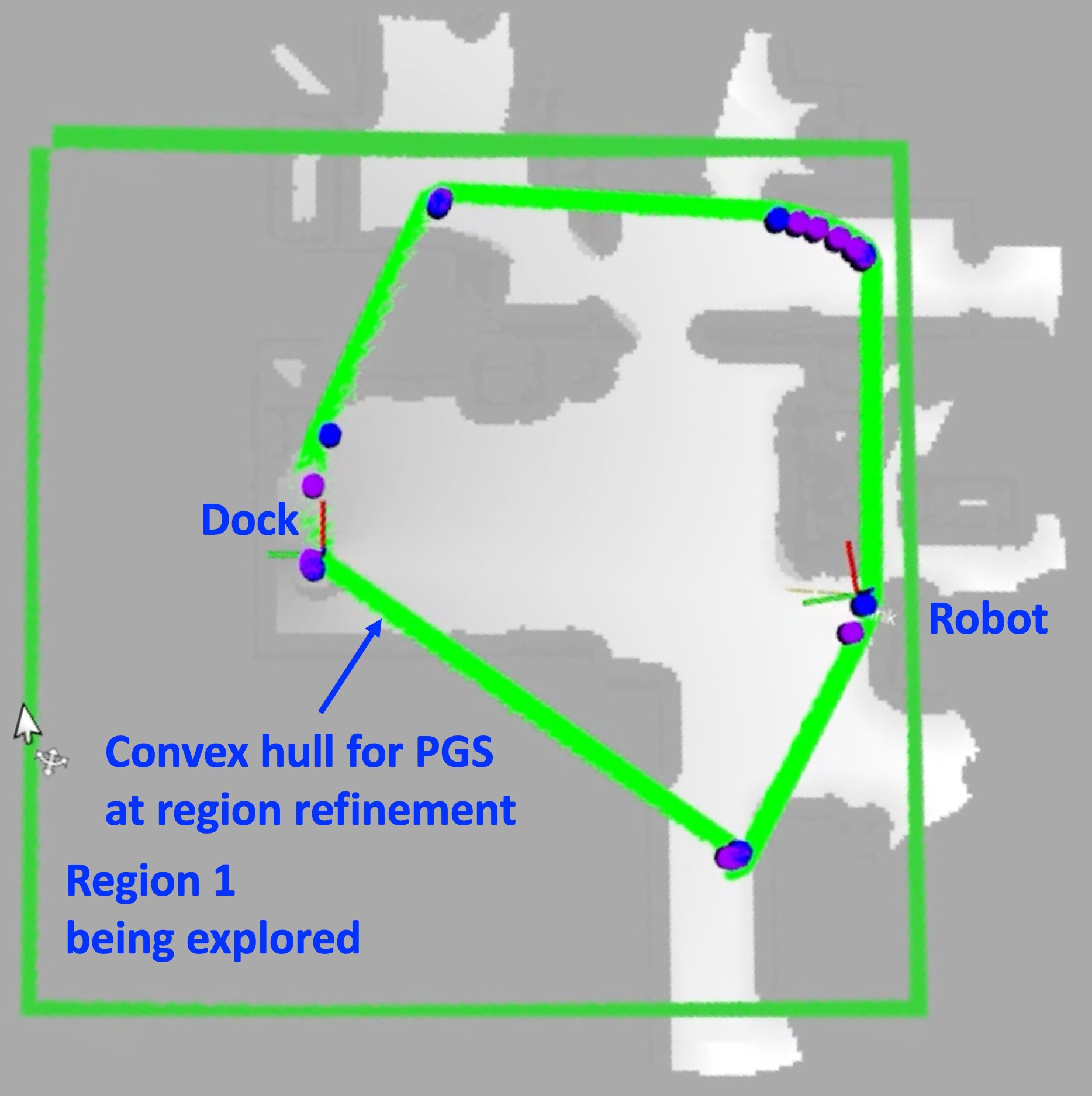}
         \caption{Region refinement of Region 1}
         \label{fig:step2}
     \end{subfigure}
     \hfill
     \begin{subfigure}[b]{0.295\textwidth}
         \centering
         \includegraphics[width=\textwidth]{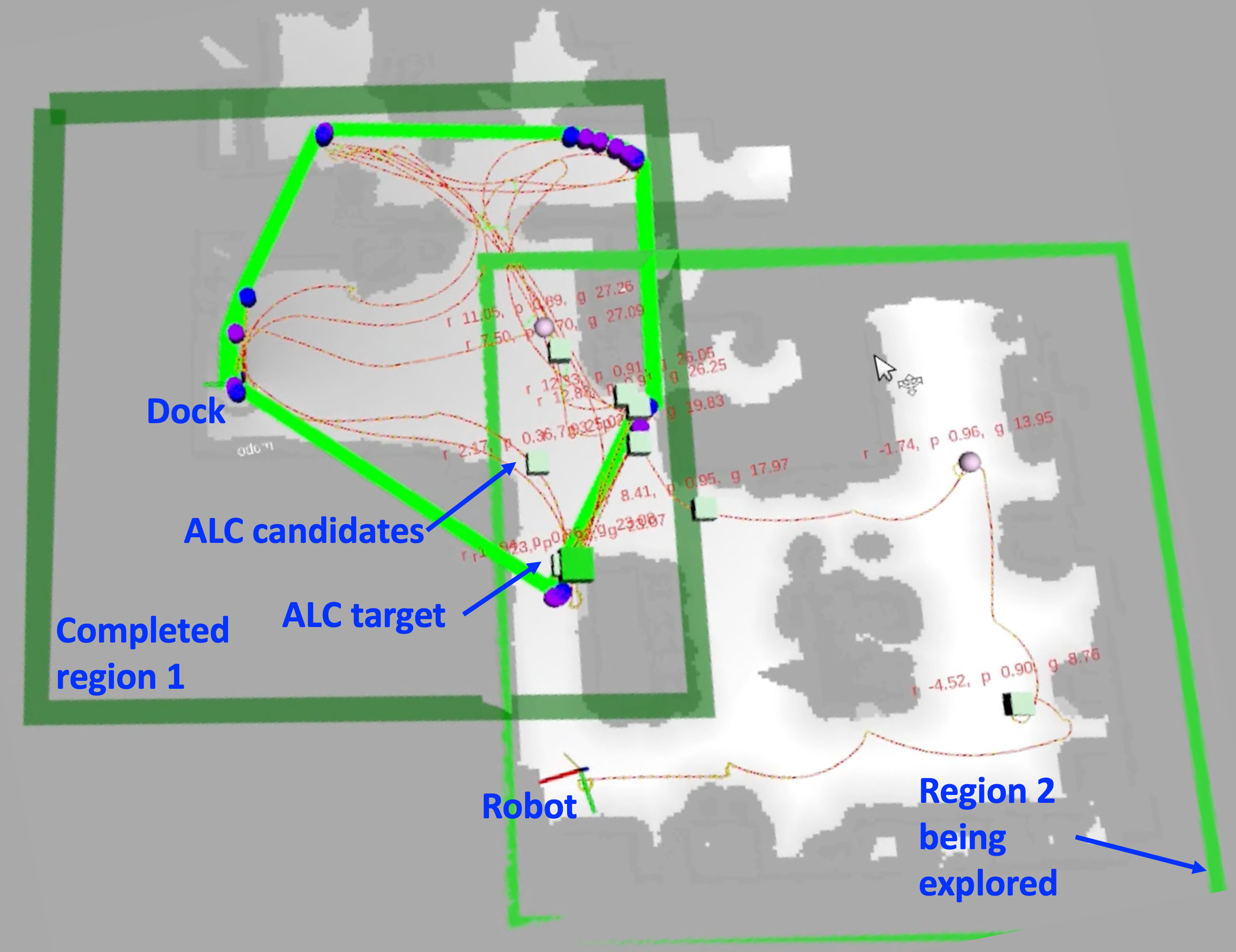}
         \caption{ALC within Region 2}
         \label{fig:step4}
     \end{subfigure}
     \hfill
     \begin{subfigure}[b]{0.245\textwidth}
         \centering
         \includegraphics[width=\textwidth]{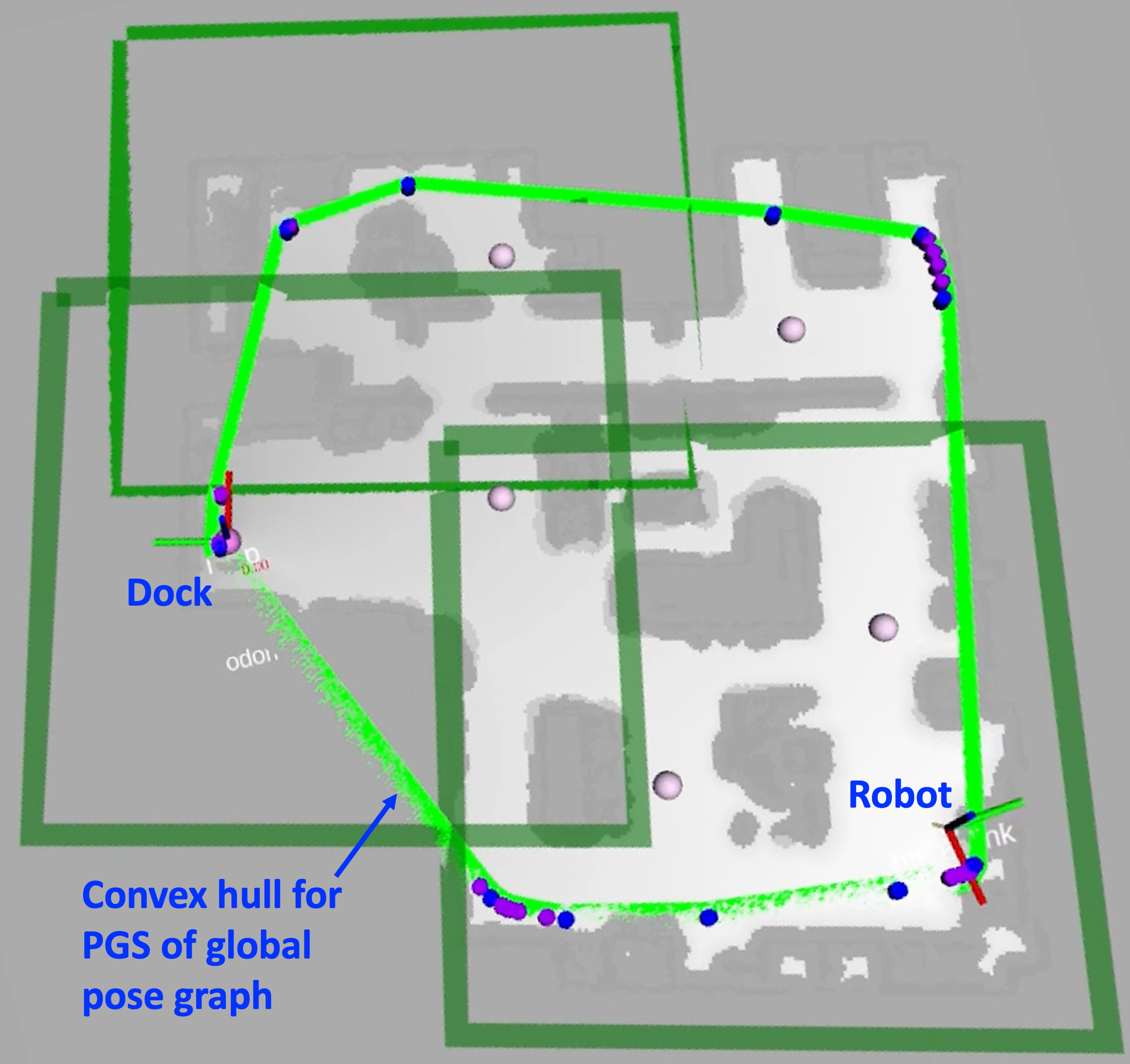}
         \caption{PGS within the global map}
         \label{fig:step5}
     \end{subfigure}
        \caption{Case study of region based exploration process simulated in HSL\#1. (a) We show Region 1 (a light green rectangle) that is currently undergoing region discovery phase. The robot focuses on addressing all frontiers within Region 1, while ignoring those outside Region 1. (b) As Region 1 enters the refinement phase, the PGS planner generates a convex hull (a light green polygon) using the keyframes of the pose graph inside the current region. Then the robot navigates along this convex hull in both clockwise and counter-clockwise directions to ensure regional pose graph stability. (c) The ALC planner decides an ALC target (a large green cube) from the set of ALC candidates (small green cubes),  all situated within Region 2. The ALC planner is activated as uncertainty reduction exceeds the threshold, and drives the robot to the ALC target to attempt a loop closure and improve pose graph local stability. (d) Upon completion of all regions (both discovery and refinement phases), the PGS planner computes a convex hull using all the keyframes on the \textit{global} pose graph. The robot then traverses all vertices of this convex hull to ensure global stability.}
        \label{fig:debug_info_viz}
\end{figure*}

\subsection{Map Segmentation}
Map segmentation into rooms is typically a post-processing step that occurs after exploration is complete and the map has been constructed. Various approaches to this task have been proposed in recent literature. In \cite{rosinol2021kimera}, a 2D slice of the 3D Euclidean Signed Distance Field (ESDF) cut right below the ceiling height is used for room segmentation. Truncating the 2D ESDF to distance above a threshold results in disconnected 2D ESDF components corresponding to individual rooms. In \cite{hughes2024foundations}, a method is presented to identify rooms by dilating the voxel map and partitioning the graph of places into disconnected components. In \cite{Taehyeon2024RoomSegment}, a large language model (LLM) is employed to further improve the segmentation performance of geometric segmentation methods. A technique is introduced in \cite{kassab2024language} to segment the pose graph and cluster its nodes using the room labels provided by Contrastive Language-Image
Pretraining (CLIP) \cite{radford2021learning}. In this paper, in contrast, our paper proposes segmenting the entire space into regions of during exploration to bootstrap the process. These regions are used to constrain exploration trajectories and facilitate the exploration process.

\subsection{SLAM-Aware Exploration Strategies}
To mitigate drift in pose estimates from a SLAM system during exploration, it is necessary to perform active loop closure (ALC) and reduce the pose graph uncertainty \cite{placed2023survey}. Loop closure is the process of recognizing a previously visited location and using this information to correct accumulated errors in a robot's estimated position and map. While pure frontier planning \cite{yamauchi1997frontier} is not incentivized to revisit previously explored areas, a dedicated ALC planner actively guides the robot to do so, thereby reducing uncertainty. Various methods have been proposed to address ALC planner design. Different metrics has been used to formulate reward/utility functions to choose the best location in a map for ALC, including expected reduction in entropy of the map \cite{gonzalez2002navigation, wu2019autonomous, stachniss2005information}, full graph A-optimality \cite{sim2005global}, and full graph D-optimality \cite{placed2022general, placed2022explorb}. Although full graph D-optimality is a powerful metrics capturing the pose graph uncertainty, it requires significant computation effort to obtain, which poses challenges for robots with limited computation power. 

In our previous work \cite{Mohit2023lighthouse}, the concept of lighthouse is introduced, which is a location in a map offering panoramic view that serves as visual reference points during exploration. The ALC planner utilizes these lighthouses, selecting the nearest ones as optimal position and periodically guiding the robot to these locations to bring down pose graph uncertainty. Later in \cite{ALCv2}, a probabilistic reward function constructed based on neighbor D-optimality is proposed to choose ALC targets, which minimizes the computation burden. This helps to maximally reduce uncertainty in the pose graph while avoid large traveling distance. This paper adapts the ALC planner proposed in \cite{ALCv2} to operate with the concept of regions and ensure the local stability of regional pose graph. This adaptation serves as one of the key building blocks of the proposed region-based exploration strategy.

\section{Notations}
Let $\bm{\mathcal{G}} = \left(\bm{\mathcal{V}}, \bm{\mathcal{E}}\right)$ be a pose graph where $\bm{\mathcal{V}} = \{\bm{v}_1, \bm{v}_2, \dots , \bm{v}_n\}$ is the set of vertices (also called graph nodes). Each graph node has a corresponding keyframe, a selected frame or pose that contains important visual or spatial information. $\bm{\mathcal{E}} = \{\bm{e}_1, \bm{e}_2, \dots, \bm{e}_m\}$ is the set of edges, each representing the relative pose measurement between the two connected keyframes. An edge in the pose graph can also be referenced as a factor. We denote the pose of the vertex $\bm{v}_i$ as $\bm{p}_{\bm{v}_i}$. We assume that the latest node $\bm{v}_r$ on the pose graph corresponds to the robot's current pose $\bm{p}_r$. Robot's $x$ and $y$ coordinates are denoted by $\texttt{p}_r$.

\section{Main Methods}

\begin{figure}[H]
  \centering
  \includegraphics[width=0.45\textwidth]{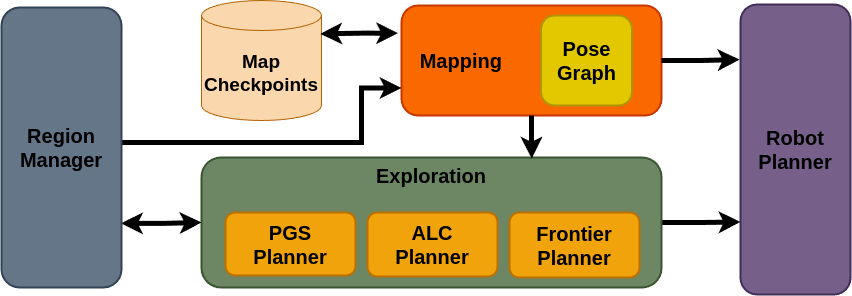}
  \caption{System architecture: The \textit{Map Checkpoints} block stores the map checkpoint after each successful explored region. The \textit{Region Manager} manages the region information. The \textit{Mapping} component handles current map management and pose graph optimization/marginalization. The \textit{Exploration} component handles the whole exploration logic using current map data and region information. This component uses three planners: 1) \textit{Frontier} planner determines the next frontier to explore, 2) \textit{Active Loop Closure} (ALC) planner performs active loop closure, 3) \textit{Pose Graph Stabilizing} (PGS) planner stabilizes the pose graph. Finally, the \textit{Robot Planner} component executes the motion to the next global plan determined by the Exploration component.}
  \label{fig:system_architecture}
\end{figure}

\subsection{System Architecture Overview}

\thisWork{} enables a mobile robot to explore the environment region by region. The overall system architecture can be seen in Figure \ref{fig:system_architecture}. At a fixed rate during each planning cycle, the Mapping component collects surrounding perceptual data to generate submaps, which are 2D occupancy grids at a smaller scale compared to the global occupancy map. Subsequently, the Mapping component aggregates these previously generated submaps into a comprehensive global occupancy map. This global map, along with an updated pose graph, is then bundled into a world snapshot. This snapshot serves as a unified representation of the environment and is made accessible to all exploration planners, ensuring consistent and up-to-date information for decision-making processes. 

The frontier planner ensures that the current region, defined by the region manager, is fully discovered. ALC planner and pose graph stabilizing planner work together to stabilize the pose graph within the region before moving to the next one. This prevents the robot from going back and forth during exploration, and improve the efficiency. The state of the map, after each successful explored region, is saved to create a map checkpoint. The exploration is finished once the whole environment is explored and globally stabilized.

As each region is explored and stabilized, the region manager informs the mapping module to perform pose graph marginalization. This process selectively removes keyframes from the completed region, leveraging the relative local stability of these keyframes. This approach significantly reduces computational load and memory requirements, enabling the system to efficiently scale to larger environments.


To enhance robustness, \thisWork{} incorporates an exploration recovery mechanism. In the event of a failure while exploring a new region, the system can leverage the latest checkpoint as a recovery point. This allows the robot to resume exploration from a known, reliable state, effectively minimizing the impact of failures and ensuring continuity of the mapping process.

Once all regions are explored, the pose graph stabilizing planner executes a global graph stabilization process to globally stabilize all regions together by smartly traversing regions as shown in light green in Figure \ref{fig:step5}. Notice that each region has moved relatively to each other, but that each region remained relatively fixed internally. A global optimization of all poses could also be performed, but this will be decided by the SLAM layer based on the amount of poses and requirements to perform the optimization.

\subsection{Region Based Exploration} \label{sec:reb}
\begin{figure}[t]
  \centering
  \includegraphics[width=0.48\textwidth]{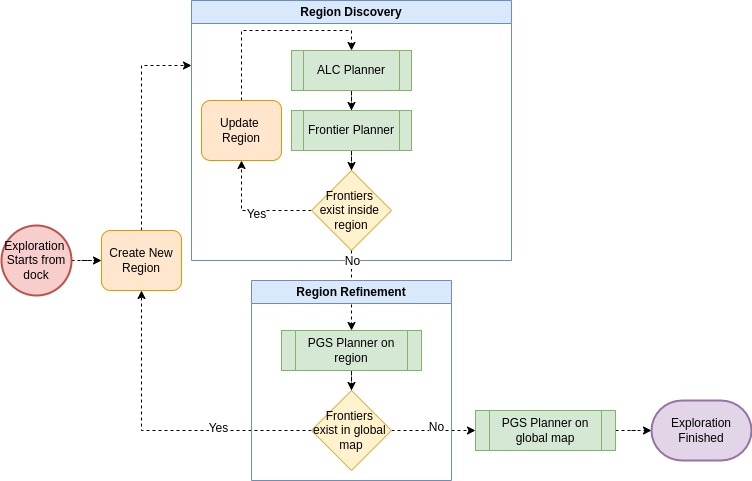}
  \caption{Region Based Exploration Flow Chart. During each exploration planning cycle at a fixed rate, we begin with the region discovery phase, alternating between ALC and frontier planners. After the region is fully discovered, we enter the region refinement phase using the PGS planner. After a region is completed (i.e. discovered and stabilized), we create a new one, and repeat the process again. When all regions are completed, we run the PGS planner on global scale to stabilize the global map, concluding the exploration.}
  \label{fig:rbe_flow_chart}
\end{figure}

\textit{Region Definition:} A region is defined as a subset of the environment's space. In this paper, we represent regions using rectangles. The size of these rectangles is adjustable. Each rectangle is initially created with relatively small dimensions, denoted as $(W_{min}, H_{min})$ for width and height, respectively. As the robot moves, the region manager gradually expands the region to encompass a small neighborhood around robot's current position $\texttt{p}_r \in \mathbb{R}^2$, where the neighborhood is defined as $\{\texttt{p} \in \mathbb{R}^2: ||\texttt{p} - \texttt{p}_r||_2 \leq r\}$, and $r > 0$ is the neighborhood radius. This expansion continues until both the width and height reach their maximum allowable sizes, represented as $(W_{max}, H_{max})$. The region constrains the exploration trajectories, and prevents the robot from going back and forth during exploration. 

\textit{Region Discovery Phase:} The exploration process (illustrated in Figure~\ref{fig:rbe_flow_chart}) begins by establishing a new region centered around the robot's initial location at the dock by the region manager, marking the start of the region discovery phase. During this phase, the robot executes a planning cycle at 1 Hz, first evaluating the triggering condition for the ALC planner. This condition is met when the expected uncertainty reduction $\Delta U$ between the robot pose $\bm{p}_r$ and the ALC target $\bm{\tau}^*$, if a loop closure constraint were to be established between them, exceeds a predetermined threshold:
\begin{align}
   \Delta U(\bm{p}_r, \bm{p}_{\bm{\tau}^*}) = \frac{\det \left(\bm{\Sigma}_-(\bm{p}_r, \bm{p}_{\bm{\tau}^*})\right)} {\det \left(\hat{\bm{\Sigma}}_{+}(\bm{p}_r, \bm{p}_{\bm{\tau}^*})\right)}, \label{eq:uncertainty_reduction_cov}
\end{align}
where $\bm{\Sigma}_{-}$ and $\hat{\bm{\Sigma}}_+$ denote prior and estimated posterior covariance matrices between node pairs before and after establishing a loop closure between them. The determinant of the covariance matrices, referred to as neighbor D-optimality, quantifies uncertainty. This representation of uncertainty reduction achieves $\mathcal{O}(1)$ complexity, which is more efficient than the graph D-optimality metrics from \cite{placed2022general} with $\mathcal{O}(n^3)$ complexity. $n$ is the number of keyframes in the pose graph. Posterior matrix estimation is detailed in \cite{ALCv2}.

When the ALC planner's triggering condition is met, it takes precedence over the frontier planner, and starts to guide the robot towards the optimal ALC target to achieve loop closure. The ALC target is chosen among a set of ALC candidates that can maximize the uncertainty reduction $\Delta U(\bm{p}_r, \bm{p}_{\bm{\tau}})$ while minimizing traveling distance. An ALC candidate $\bm{\tau}$ is a cluster of neighboring keyframes with abundant visual features for loop closures. The way of constructing ALC candidates can be found in \cite{ALCv2}. Nonetheless, in this work, the ALC target is selected from candidates exclusively within the current region. In summary, at each planning cycle, the ALC planner computes an ALC target among the candidates, and will be activated if the uncertainty reduction at the target is above the threshold.

If the ALC condition is not satisfied, the robot activates the frontier planner. Frontiers are contiguous unknown cells adjacent to free cells \cite{yamauchi1997frontier}. The planner ignores frontiers outside the current region and calculates costs for remaining frontiers using:
\begin{align}
C(F_i) = C_L(\texttt{p}_r, \texttt{p}_{F_i}, \bm{\mathcal{M}}) + & \beta_1 C_A(\texttt{p}_r, \texttt{p}_{F_i}) + \nonumber \\
&\beta_3 C_S(F_i, \hat{F}) - \beta_4 G(F_i) \label{eq:front_cost}
\end{align}
where $C_L(\texttt{p}_r, \texttt{p}_{F_i}, \bm{\mathcal{M}})$ represents the shortest distance from the robot position $\texttt{p}_r$ to the frontier position $\texttt{p}_{F_i}$ using the A$^*$ algorithm on the occupancy map $\bm{\mathcal{M}}$. $C_A(\texttt{p}_r, \texttt{p}_{F_i})$ is angular penalty for non-holonomic robots to avoid choosing frontiers that require too much turning. The switch cost $C_S(F_i, \hat{F})$ penalize selecting a frontier $F_i$ in this planning cycle that differs from the previously chosen frontier $\hat{F}$. This discourages frequent changes in exploration targets. The last term $G(F_i)$ measures the information gain, which is approximated using the number of cells making up a frontier $F_i$. The weights $\beta_i$ balance travel distance, robot constraints, exploration consistency, and potential information gain. At each planning cycle, the frontier planner re-evaluates all frontiers within the region using the cost function (equation~\ref{eq:front_cost}). It employs a greedy approach, directing the robot to the frontier with the lowest cost until there is no more frontier in the current region. 

\textit{Region Refinement Phase:} Upon exploring all frontiers within the current region, the process transitions to the region refinement phase. This phase aims to enhance the stability of the regional pose graph. While the ALC planner's loop closure constraints ensure local stability by connecting neighboring nodes, the overall pose graph within the region still benefits from additional stabilization measures.

During the region refinement phase, the robot activates the Pose Graph Stabilizing (PGS) planner. This planner aims to enhance stability by adding more constraints to the periphery of the \textit{regional} pose graph. The process begins by identifying the keyframes of the regional pose graph and computing their convex hull using the QuickHull algorithm \cite{barber1996quickhull}.

The PGS planner then guides the robot to traverse each vertex of this convex hull, both in clockwise and counterclockwise directions. This systematic movement allows the robot to establish new loop closure constraints, effectively connecting the vertices and reinforcing the overall stability of the pose graph within the region. This serves as the prerequisite step for the regional pose graph marginalization mechanism detailed in Section~\ref{sec:pose_graph_marginalization}.

\textit{Global Map Stabilization:} Upon completion of the region refinement phase, if no frontiers remain within the global map, we consider the global map fully explored. At this point, the system transitions to the PGS planner, extending its scope to the entire global map. Specifically, it navigates the robot along the convex hull formed by the keyframes of the \textit{global} pose graph. This process aims to reinforce the overall stability of the map by establishing additional loop closure constraints on a global scale. During this process, while the regions may shift relative to one another, the internal structure of each region remains largely intact.

The exploration process concludes once this global stabilization phase is complete. This final step ensures that the entire mapped area benefits from enhanced pose graph stability, resulting in a more accurate and reliable representation of the explored environment.

The entire region based exploration strategy, simulated in a home environment of size $150$ m$^2$ with photorealistic Unreal simulation Engine, is showcased in Figure~\ref{fig:debug_info_viz}. The configuration of the robot used in simulation can be found in Section~\ref{sec:ex_setup}.

\subsection{Pose Graph Marginalization}\label{sec:pose_graph_marginalization}
\begin{figure}[t]
  \centering
  \includegraphics[width=0.5\textwidth]{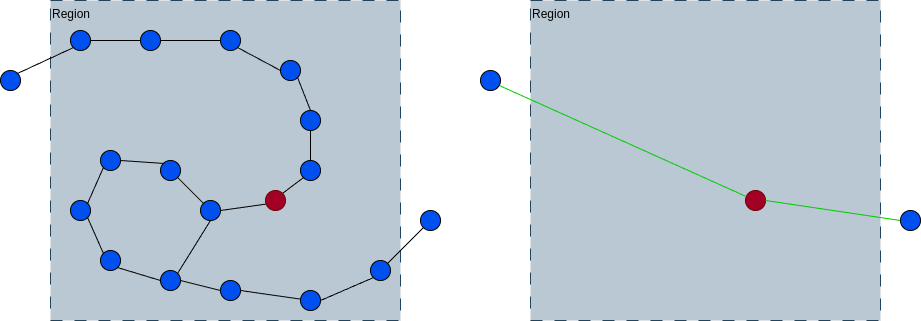}
  \caption{Pose Graph Marginalization over a Region. The left image shows the pose graph after exploring and stabilizing a region (grey box), with keyframes represented as blue and red dots and factors as black edges. A subset of keyframes (of size 1 for demonstration purposes here), represented as a red dot, is selected within the region to perform pose graph marginalization. The marginalization process removes all blue keyframes while retaining the information of the factors in new connections. The right figure displays the pose graph after marginalization, which includes only the single anchor keyframe within the region and new factors (green edges) connecting the remaining keyframe to keyframes outside of the region. This process significantly reduces the complexity of the pose graph while preserving essential spatial information.}
  \label{fig:pose_graph_marginalization}
\end{figure}

In the context of region-based exploration, the pose graph $\bm{\mathcal{G}}$ can be segmented into one or multiple pose graph $\bm{\mathcal{G}_i}=\{\bm{\mathcal{V}_i}, \bm{\mathcal{E}_i}\}$ for each region $i$, along with $\bm{\mathcal{G}_{out}}=\{\bm{\mathcal{V}_{out}}, \bm{\mathcal{E}_{out}}\}$ which represents the set of keyframes and edges that are not contained within any specific region, such as those in transitional areas:
\begin{align}
\small
\bm{\mathcal{G}} = \left(\big\{\bm{\mathcal{G}_1}, \bm{\mathcal{G}_2} \dots \bm{\mathcal{G}_n} \big\}, \bm{\mathcal{G}_{out}} \right)
\end{align}
\begin{align}
\small
\bm{\mathcal{G}} = \left(\big\{\{\bm{\mathcal{V}_1}, \bm{\mathcal{E}_1}\}, \{\bm{\mathcal{V}_2}, \bm{\mathcal{E}_2}\} .. \{\bm{\mathcal{V}_n}, \bm{\mathcal{E}_n}\}\big\}, \{\bm{\mathcal{V}_{out}}, \bm{\mathcal{E}_{out}}\}\right),
\end{align}

As the robot explores, loop closure detection is performed to identify revisited locations. When a loop closure is detected, a new constraint is added to the pose graph, connecting the current pose to a previously visited pose. This process helps correct for accumulated drift in the trajectory. After each new loop closure, we optimize the pose graph using nonlinear techniques. Specifically, we employ the Gauss-Newton method with Cholesky decomposition to iteratively solve the resulting sparse system of equations until convergence \cite{kummerle2011g2o}. For a pose graph with $n$ keyframes, each iteration has a time complexity of $\mathcal{O}(n^3)$ and a memory complexity of $\mathcal{O}(n^2)$, in the worst case, when using Cholesky decomposition. This highlight the critical need to manage and reduce the number of keyframes in the pose graph, especially for large-scale environments.

To address this need, our approach implements pose graph marginalization as a key component of our region-based exploration strategy. This process is designed to optimize computational resources and memory usage while maintaining map integrity. It involves selectively removing keyframes (nodes) from the pose graph while preserving the essential information contained in the relative poses (factors) connected to these nodes. By creating new factors that capture the information from the original connections, we can reduce the size of the graph while maintaining its overall structure and information content.

Our approach implements a specific marginalization policy tied to the exploration and stabilization status of each region. Before considering any keyframes for marginalization, we first verify that the current region has been fully explored and stabilized. This ensures that we only simplify areas of the map that have been comprehensively mapped and have a high degree of certainty. Once a region is confirmed to be stable, we proceed to marginalize most keyframes within it, retaining only a subset to serve as anchor points for the region. These retained keyframes are selected based on a policy that considers criteria such as spatial distribution and feature richness. This subset of anchors is vital for maintaining global consistency of the map between regions while still allowing for significant reduction of keyframes in the pose graph, which is key to reduce compute and memory requirements as mentioned previously.

Taking region 1 as example, the pose graph $\bm{\mathcal{G}_1}$, before marginalization, is defined as:
\begin{align}
\small
\bm{\mathcal{G}_1} = \{\bm{\mathcal{V}_1}, \bm{\mathcal{E}_1}\} = \big\{\{\bm{v}_{1_1}, \bm{v}_{1_2},..\bm{v}_{1_n}\}, \{\bm{e}_{1_1}, \bm{e}_{1_2},.. \bm{e}_{1_n}\}\big\},
\end{align}
where $v_{1_i}$ and $e_{1_i}$ are respectively the keyframes and factors within region 1. After marginalization, the pose graph can now be represented as:
\begin{align}
\small
\bm{\mathcal{G}_1} = \big\{\bm{\mathcal{V}'_1}, \bm{\mathcal{E}'_1}\} = \{\{\bm{v}^{a}_{1}, \dots \bm{v}^{a}_{n}\}, \{\bm{e}'_{1_1}, \bm{e}'_{1_2},... \bm{e}'_{1_n}\}\big\},
\end{align}
where $\bm{v}^a_{i}$ are the subset of remaining keyframes representing the region anchors, and $\bm{e_i}'$ are the new factors retaining the information of the marginalized factors.

Different methods can be used such as \cite{mladen2016nfr,carlevaris2014generic} to estimate these new factors. For this work we use the Nonlinear Factor Recovery (NFR) method as in \cite{mladen2016nfr}. We first compute the target information matrix over the elimination clique by marginalizing out the keyframes to be removed using the Schur complement, which gives us the marginalized normal distribution target $P(x)$. We then solve a convex optimization problem to recover a set of nonlinear factors such that the resulting linearized distribution $Q(x)$ minimizes the Kullback-Leibler divergence (KLD) with respect to $P(x)$ over the same sample space $\mathcal{X}$:
\begin{align}
D_{KL}(P || Q) = \sum_{x \in \mathcal{X}} P(x) \log \left(\frac{P(x)}{Q(x)}\right)
\end{align}

The outcome of this process, as illustrated in Figure \ref{fig:pose_graph_marginalization}, is a restructured pose graph that maintains the map's integrity while significantly reducing its complexity. The marginalization creates new nonlinear factors (represented by green edges) that encapsulate the information from all the original factors (black edges) connected to the marginalized keyframes (blue dots). To complete the process, we update the pose graph structure by removing the marginalized keyframes from the active graph and incorporating these newly created nonlinear factors. These factors are connected to the remaining keyframes in the elimination clique, including both the retained keyframe within the region and any keyframes outside the region that were originally linked to the marginalized ones. This approach preserves essential spatial relationships while eliminating redundant nodes, ensuring the pose graph remains informative yet computationally efficient. Consequently, this enables the system to scale effectively to larger environments without compromising on map quality or navigation capabilities.

\begin{figure*}[t]
     \centering
     \includegraphics[width=\textwidth]{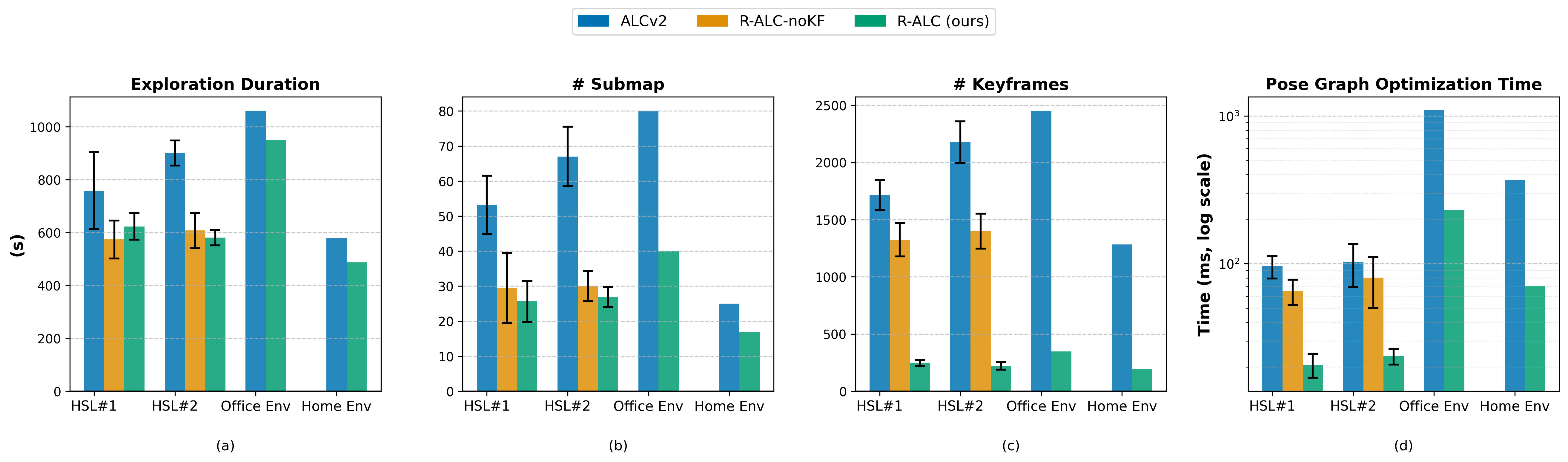}     
    \caption{Comparison of exploration duration (a), submap count (b), keyframe count (c), and average pose graph optimization time (d) between \thisWork{}, \thisWork{} without keyframe marginalization, and the baseline \alcv{}\cite{ALCv2} across different simulation environments (HSL\#1 and HSL\#2) and real-world environments (Home and Office Env). Lower values are better for all plots. \thisWork{} reduces exploration duration (by $10\%$ in office, $15\%$ in home) and generates a global occupancy map with fewer submaps ($50\%$ fewer in office, $32\%$ fewer in home). Additionally, \thisWork{} uses significantly fewer keyframes ($85\%$ fewer) to construct the SLAM pose graph, resulting in a substantial decrease ($78-80\%$) in the average pose graph optimization time.}
    \label{fig:exploration_duration_and_mapping_overhead}
\end{figure*}

\subsection{Exploration Recovery} \label{sec:exploration_recovery}
Exploration failures due to map distortions can occur when localization drift or sensing errors prevent the device from planning paths and navigating to previously visited locations. 
Restarting exploration from scratch is time consuming especially as the map size grows. To mitigage the cost of handling such failures \thisWork{} saves snapshots of the map -- capturing both the SLAM pose graph and the navigation occupancy map -- at the end of each region once the map has stabilized. In the event of such a failure, an exploration recovery behavior is triggered: the device either autonomously navigates back to the starting location or is manually returned if navigation is not feasible. Once there, the snapshot corresponding to the latest completed region is loaded, allowing region-based exploration to resume from a stable state.

\section{Experimental Results} \label{sec:experimental_results}


In this section, we address the following questions: 1) How much improvement does \thisWork{} achieve over a SOTA SLAM-aware exploration algorithm in terms of exploration duration, as well as the number of occupancy submaps and keyframes required to represent the navigation and SLAM maps?
2) How effective is \thisWork{}’s region-based keyframe marginalization component in reducing the size of the SLAM pose graph, thereby lowering the memory and computational resources needed for storage and processing?
3) How accurate are the maps generated during autonomous exploration using \thisWork{}?

\subsection{Experimental Setup} \label{sec:ex_setup}
We evaluate \thisWork{} using a differential-drive home robot equipped with an IMU, a stereo RGB camera pair, and a depth camera. Localization is performed using a visual SLAM algorithm that integrates stereo camera images, IMU readings, and wheel encoder data, while the depth camera readings are used to construct a 2D occupancy map of the environment.
Experiments are conducted in both real-world and simulated environments. The real-world tests take place in a~\SI{117}{\square\meter} home and a~\SI{210}{\square\meter} office. For controlled and repeatable evaluation, we also use a photorealistic Unreal Engine-based simulation environment with two distinct home environments of size~\SI{204}{\square\meter} and~\SI{150}{\square\meter}.

\begin{figure*}[t]
     \centering
     \includegraphics[width=0.95\textwidth]{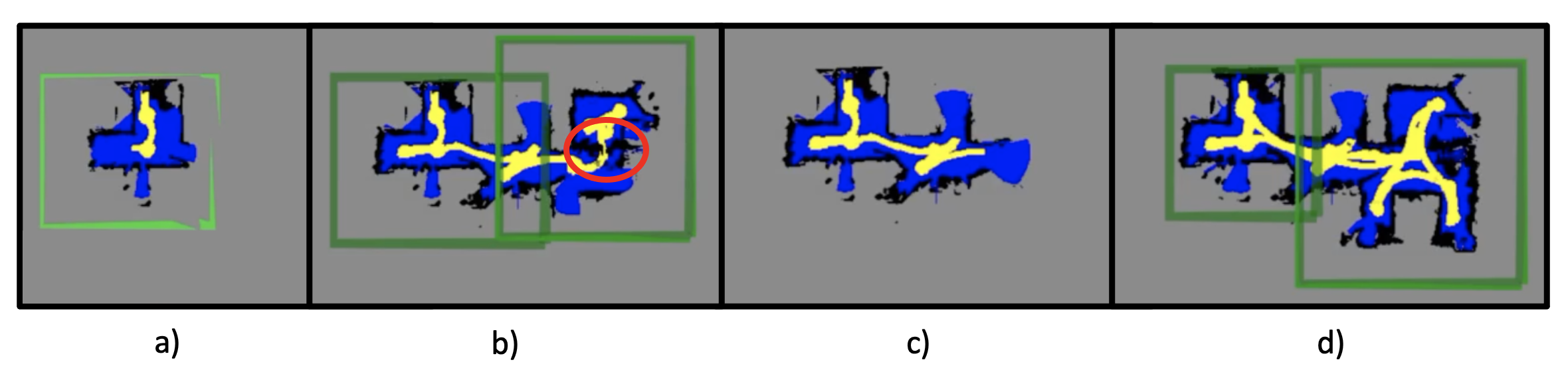}     
    \caption{Example of an exploration recovery scenario from a real device test. The figures visualize the global occupancy map of the environment as it is being generated during an exploration session. Black cells indicate occupied areas, blue represents free space, and yellow shows the robot's trajectory (dilated by its radius). Green rectangles denote exploration regions. a) The device begins exploration. b) It encounters a failure while exploring the second region due to inability to find a path to a previously visited location. The highlighted area is blocked by simulated clutter, representing potential map distortions caused by false obstacle detection and/or localization drift. c) The device is returned to its dock, and exploration is restarted. The system restores the latest saved progress, corresponding to the map state at the completion of exploration of the first region. d) The device explores the remaining unvisited areas and completes the exploration task.}
    \label{fig:exploration_recovery_eg}
\end{figure*}

\subsection{Exploration Speed and Mapping Overhead}

We compare \thisWork{} against our previous work, \alcv{}~\cite{ALCv2}, and an ablated version of \thisWork{}, where the keyframe marginalization component is disabled, evaluating their performance in terms of exploration speed and mapping overhead. Figure~\ref{fig:exploration_duration_and_mapping_overhead} presents the results from autonomous exploration trials in simulation environments (5 runs for each algorithm-environment pair) as well as single-trial exploration runs in the real world.
\thisWork{} significantly reduces exploration time and generates maps covering the same space using substantially fewer occupancy submaps and pose graph keyframes. The number of submaps and keyframes directly correlates with map size and, consequently, the SLAM algorithm's memory consumption. \thisWork{} without keyframe marginalization achieves similar improvements to \thisWork{} in reducing the number of submaps; however, the resultant SLAM pose graph includes a significantly larger number of keyframes, as expected. The region-based exploration approach reduces redundant revisits, leading to fewer submaps, while keyframe marginalization further optimizes memory usage, and reduces the computational overhead of pose graph optimization by eliminating redundant keyframes within the same region. The improvement in average pose graph optimization time is particularly significant during real robot deployment, where computational resources are limited. On the ARM processor of the robot, solving the pose graph optimization problem takes an order of magnitude longer than on a laptop for maps of the same size, making the observed speedup especially impactful. As shown in the results for the Office environment in Figure~\ref{fig:exploration_duration_and_mapping_overhead}, the average optimization time of the pose graph—which is executed frequently, particularly when a new constraint is added—is reduced by approximately 80\%, from~\SI{1.1}{\second} to~\SI{0.2}{\second}.

\begin{figure}[t]
  \centering
  \includegraphics[width=1.0\columnwidth]{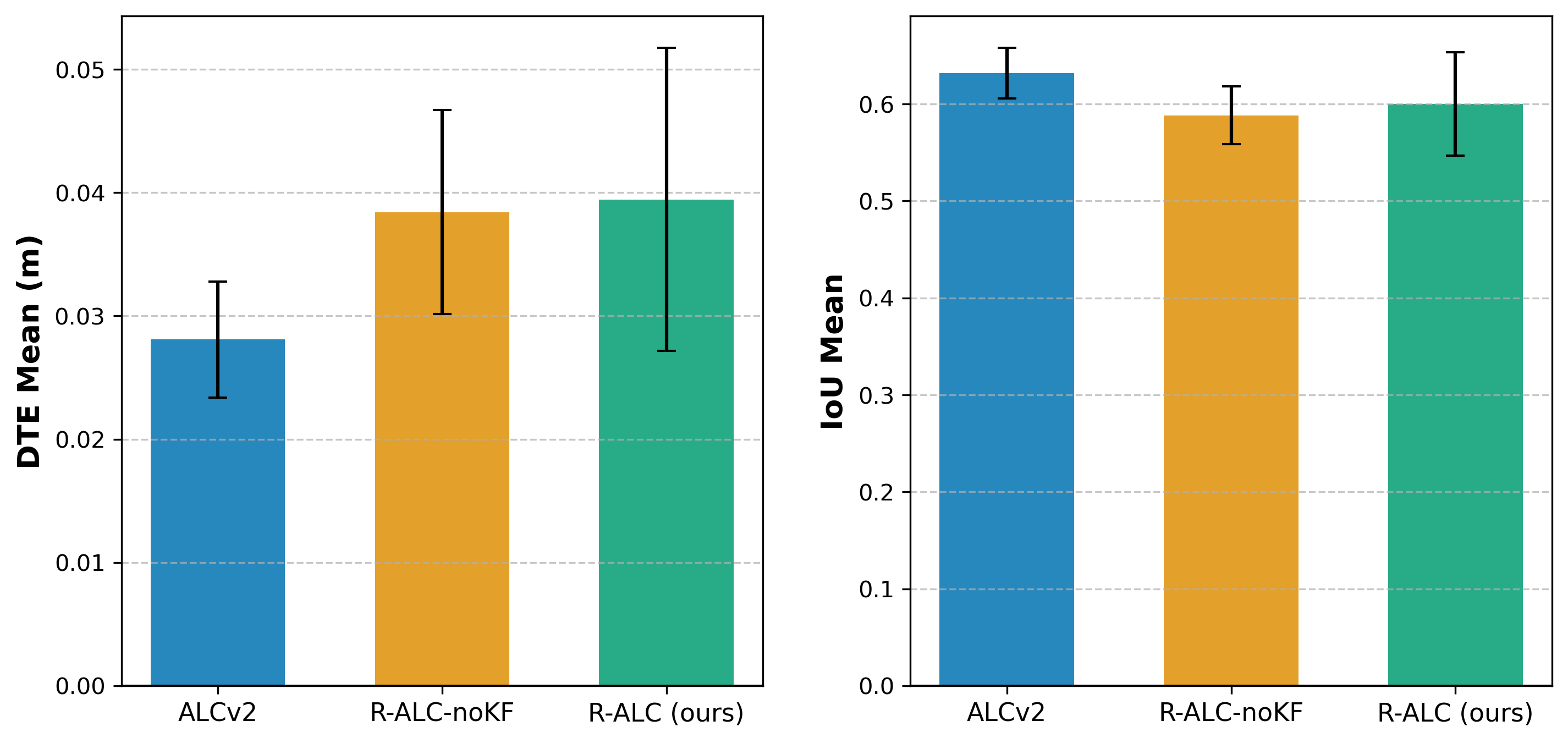}
  \caption{Comparison of the quality of occupancy maps generated by \thisWork{} and the baseline exploration algorithms, evaluated using the aggregate mean IoU (higher is better) and mean DTE (lower is better) across all exploration runs in two simulation environments. The IoU results show comparable alignment between the reference and test occupancy maps for all three exploration methods. The DTE metric results show that the occupancy map errors between the exploration methods differ by less than ~\SI{2}{\centi\meter}, which is smaller than the grid cell size threshold of ~\SI{5}{\centi\meter}.}
  \label{fig:map_quality_results_quantitative}
\end{figure}

\begin{figure}[t]
\centering
 \begin{subfigure}[b]{0.45\linewidth}
  \includegraphics[width=1.0\linewidth, trim=120 90 110 190,clip]{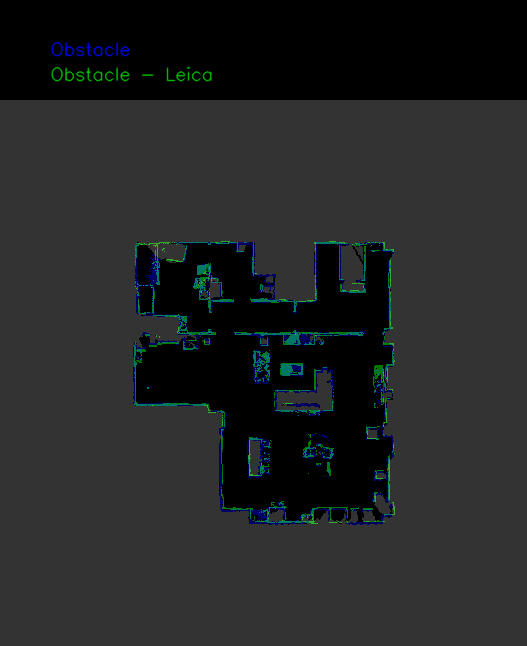}
 	\caption{HSL\#1}
 	\label{fig:hsl_1}
 \end{subfigure}
 \begin{subfigure}[b]{0.45\linewidth}
  \includegraphics[width=1.0\linewidth, trim=90 110 110 210,clip]{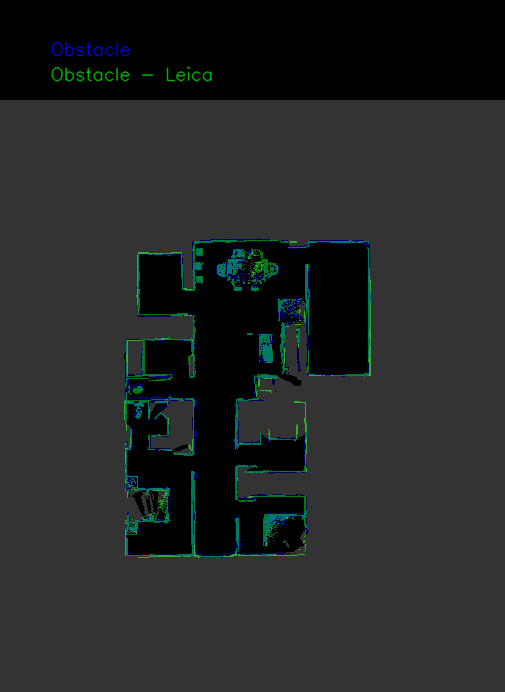}
 	\caption{HSL\#2}
 	\label{fig:hsl_2}
 \end{subfigure} 
  \caption{Visualization of the occupancy maps generated after autonomous exploration with \thisWork{} overlaid on the ground truth maps for the \subref{fig:hsl_1} HSL\#1 and \subref{fig:hsl_2} HSL\#2 simulation environments. The occupied cells in the test and reference maps are visualized in blue and green, respectively. The generated maps are globally consistent with the reference maps.}
 \label{fig:map_quality_results_qualitative}
\end{figure}

\subsection{Exploration Map Quality}
To evaluate the quality of the generated occupancy maps, we compare them against reference maps generated by running exploration using ground truth location information from the simulation. We compute two metrics to quantify the similarity between the test-time maps and the reference maps.

\paragraph{Intersection over Union}
Intersection over Union (IoU) measures the overlap between two maps relative to their combined area for each semantic class: occupied and free space. Let $M$ be the generated grid map and $M_{\text{gt}}$ be the ground truth grid map. Both maps are aligned and cropped to the same size and resolution. The IoU for a given class $c$ is computed as:
\begin{equation}
\text{IoU}_c  =  \frac{\sum_{\langle u, v \rangle \in \mathcal{D}}{\left[ M_{\text{gt}} \left(u, v \right) = c  \land M\left(u, v \right) = c\right] }}{\sum_{\langle u, v \rangle \in \mathcal{D}}{\left[ M_{\text{gt}} \left(u, v \right) = c  \lor M\left(u, v \right) = c\right]}}
\end{equation}
where $ c \in \left\{ c_i \right\}_{i=1:C+1} = \left\{\text{occupied}, \text{unoccupied}, \text{unknown} \right\}$ belongs to the set of semantic classes. Also, $\mathcal{D} = \left\{\langle u, v \rangle  |   M_{\text{gt}}\left(u, v \right) \neq \text{unknown} \right\}$ is the set of grid cell locations where the ground truth labels are known. An IoU value of 1 indicates a perfect match for class $c$, while an IoU of 0 indicates no overlap. To obtain an aggregate similarity score across all classes, we compute the mean IoU $\text{mIoU} = \frac{1}{C} \sum_c{\text{IoU}_c}$.

\paragraph{Distance Transform Error}
The Distance Transform Error (DTE) quantifies the discrepancy between a generated grid map and the ground truth by comparing the distance of each cell to the nearest cell of a given semantic class (e.g., occupied) in both maps. This metric accounts for spatial misalignments, offering a more nuanced evaluation than IoU. For each semantic class $c$, we define the DTE as:
\begin{equation}
    \text{DTE}_{c} = \frac{1}{|\mathcal{D}|}\sum_{\langle u , v \rangle \in \mathcal{D}} | T_c \left(u, v \right) - T_{c_{\text{gt}}}\left(u, v \right)|  \quad ,
\end{equation}
where $T_c \left(u, v \right)$ is the distance transform of class $c$ in the generated map $M$, representing the distance from cell $\langle u, v \rangle$ to the nearest cell labeled $c$, and $T_{c_{\text{gt}}}$ is the corresponding distance transform computed from the ground truth map $M_{\text{gt}}$. To obtain an aggregate measure across all semantic classes, we compute the mean DTE (mDTE) $= \frac{1}{C} \sum_c{\text{DTE}_c}$.

Figure~\ref{fig:map_quality_results_quantitative} presents the aggregate mDTE and mIoU results for maps generated by \thisWork{} and the baseline methods across exploration trials in the two simulation environments. \thisWork{} achieves an IoU score comparable to that of the baseline (\alcv{} \cite{ALCv2}). The mDTE metric, which captures finer map errors, is slightly higher for \thisWork{}. However, the difference is less than ~\SI{2}{\centi\meter} -- smaller than the~\SI{5}{\centi\meter} cell width used in the occupancy maps --making the impact negligible in practice. Figure~\ref{fig:map_quality_results_qualitative} provides qualitative visualizations of sample maps generated by \thisWork{}, overlaid on the ground truth maps, showing that the generated maps remain globally consistent with the reference maps. \thisWork{} maintains this level of mapping accuracy while significantly reducing exploration time, thanks to its ability to generate more compact navigation maps and a smaller pose graph, which in turn lowers memory consumption.


\subsection{Qualitative Results}
As explained in Section~\ref{sec:exploration_recovery}, \thisWork{} supports exploration recovery by creating map snapshots at the end of each exploration region. Figure~\ref{fig:exploration_recovery_eg} presents an example from a real robot experiment, illustrating how \thisWork{} restores saved map snapshots to prevent restarting exploration from scratch in the event of failures.

\section{CONCLUSION AND FUTURE WORKS}

The paper introduces a region based approach for
autonomous exploration to improve map quality and efficiently explore the environment. 
By exploring the entire space region by region we avoid revisiting the area multiple times. Also
our approach to stabilize a region before moving to the next one ensures recovering from a stable map
in case of failures. 

As future work
\begin{inparaenum}
    \item We would like to integrate semantics to define a region.
    \item There are areas in the map which have low visual feature coverage, hence a low chance of relocalization. We would like to improve the exploration algorithm to maximize the observed distinct visual features as well.
    \item Currently, we perform a global stabilization step at the end of exploration to improve the global map consistency. We would like to extend this further for better handling of regions with identified map distortions. 
    \item Our marginalization approach could extend beyond keyframes to include submaps within each region. By combining multiple submaps attached to keyframes in the same region, we could create a single compact representation. We would like to implement this submap-level marginalization to further reduce compute and memory.
\end{inparaenum} 

\section*{ACKNOWLEDGMENT}
The authors would like to thank Jong Jin Park, Rajasimman Madhivanan, Harry Cheng, and Arnie Sen for sharing their ideas that went into this work and for supporting the delivery of this work.

\bibliographystyle{IEEEtran}
\bibliography{refs}

\begin{thebibliography}{10}
\providecommand{\url}[1]{#1}
\csname url@samestyle\endcsname
\providecommand{\newblock}{\relax}
\providecommand{\bibinfo}[2]{#2}
\providecommand{\BIBentrySTDinterwordspacing}{\spaceskip=0pt\relax}
\providecommand{\BIBentryALTinterwordstretchfactor}{4}
\providecommand{\BIBentryALTinterwordspacing}{\spaceskip=\fontdimen2\font plus
\BIBentryALTinterwordstretchfactor\fontdimen3\font minus
  \fontdimen4\font\relax}
\providecommand{\BIBforeignlanguage}[2]{{%
\expandafter\ifx\csname l@#1\endcsname\relax
\typeout{** WARNING: IEEEtran.bst: No hyphenation pattern has been}%
\typeout{** loaded for the language `#1'. Using the pattern for}%
\typeout{** the default language instead.}%
\else
\language=\csname l@#1\endcsname
\fi
#2}}
\providecommand{\BIBdecl}{\relax}
\BIBdecl

\bibitem{bircher2016receding}
A.~Bircher, M.~Kamel, K.~Alexis, H.~Oleynikova, and R.~Siegwart, ``Receding
  horizon" next-best-view" planner for 3d exploration,'' in \emph{2016 IEEE
  international conference on robotics and automation (ICRA)}.\hskip 1em plus
  0.5em minus 0.4em\relax IEEE, 2016, pp. 1462--1468.

\bibitem{dang2019graph}
T.~Dang, F.~Mascarich, S.~Khattak, C.~Papachristos, and K.~Alexis,
  ``Graph-based path planning for autonomous robotic exploration in
  subterranean environments,'' in \emph{2019 IEEE/RSJ International Conference
  on Intelligent Robots and Systems (IROS)}.\hskip 1em plus 0.5em minus
  0.4em\relax IEEE, 2019, pp. 3105--3112.

\bibitem{ALCv2}
H.~Yin, J.~J. Park, M.~Almeida, M.~Labrie, J.~Zamiska, and R.~Kim,
  ``Probabilistic active loop closure for autonomous exploration,'' in
  \emph{2024 IEEE International Conference on Robotics and Automation (ICRA)},
  2024, pp. 18\,048--18\,054.

\bibitem{Mohit2023lighthouse}
M.~Deshpande, R.~Kim, D.~Kumar, J.~J. Park, and J.~Zamiska, ``Lighthouses and
  global graph stabilization: Active slam for low-compute, narrow-fov robots,''
  in \emph{2023 IEEE international conference on robotics and automation
  (ICRA)}, 2023.

\bibitem{rosinol2021kimera}
A.~Rosinol, A.~Violette, M.~Abate, N.~Hughes, Y.~Chang, J.~Shi, A.~Gupta, and
  L.~Carlone, ``Kimera: From slam to spatial perception with 3d dynamic scene
  graphs,'' \emph{The International Journal of Robotics Research}, vol.~40, no.
  12-14, pp. 1510--1546, 2021.

\bibitem{hughes2024foundations}
N.~Hughes, Y.~Chang, S.~Hu, R.~Talak, R.~Abdulhai, J.~Strader, and L.~Carlone,
  ``Foundations of spatial perception for robotics: Hierarchical
  representations and real-time systems,'' \emph{The International Journal of
  Robotics Research}, p. 02783649241229725, 2024.

\bibitem{Taehyeon2024RoomSegment}
T.~Kim and B.-C. Min, ``Semantic layering in room segmentation via llms,'' in
  \emph{2024 IEEE/RSJ International Conference on Intelligent Robots and
  Systems (IROS)}, 2024, pp. 9831--9838.

\bibitem{kassab2024language}
C.~Kassab, M.~Mattamala, L.~Zhang, and M.~Fallon, ``Language-extended indoor
  slam (lexis): A versatile system for real-time visual scene understanding,''
  in \emph{2024 IEEE International Conference on Robotics and Automation
  (ICRA)}.\hskip 1em plus 0.5em minus 0.4em\relax IEEE, 2024, pp.
  15\,988--15\,994.

\bibitem{radford2021learning}
A.~Radford, J.~W. Kim, C.~Hallacy, A.~Ramesh, G.~Goh, S.~Agarwal, G.~Sastry,
  A.~Askell, P.~Mishkin, J.~Clark \emph{et~al.}, ``Learning transferable visual
  models from natural language supervision,'' in \emph{International conference
  on machine learning}.\hskip 1em plus 0.5em minus 0.4em\relax PMLR, 2021, pp.
  8748--8763.

\bibitem{placed2023survey}
J.~A. Placed, J.~Strader, H.~Carrillo, N.~Atanasov, V.~Indelman, L.~Carlone,
  and J.~A. Castellanos, ``A survey on active simultaneous localization and
  mapping: State of the art and new frontiers,'' \emph{IEEE Transactions on
  Robotics}, 2023.

\bibitem{yamauchi1997frontier}
B.~Yamauchi, ``A frontier-based approach for autonomous exploration,'' in
  \emph{Proceedings 1997 IEEE International Symposium on Computational
  Intelligence in Robotics and Automation CIRA'97.'Towards New Computational
  Principles for Robotics and Automation'}.\hskip 1em plus 0.5em minus
  0.4em\relax IEEE, 1997, pp. 146--151.

\bibitem{gonzalez2002navigation}
H.~H. Gonz{\'a}lez-Banos and J.-C. Latombe, ``Navigation strategies for
  exploring indoor environments,'' \emph{The International Journal of Robotics
  Research}, vol.~21, no. 10-11, pp. 829--848, 2002.

\bibitem{wu2019autonomous}
C.-Y. Wu and H.-Y. Lin, ``Autonomous mobile robot exploration in unknown indoor
  environments based on rapidly-exploring random tree,'' in \emph{2019 IEEE
  International Conference on Industrial Technology (ICIT)}.\hskip 1em plus
  0.5em minus 0.4em\relax IEEE, 2019, pp. 1345--1350.

\bibitem{stachniss2005information}
C.~Stachniss, G.~Grisetti, and W.~Burgard, ``Information gain-based exploration
  using rao-blackwellized particle filters.'' in \emph{Robotics: Science and
  systems}, vol.~2, 2005, pp. 65--72.

\bibitem{sim2005global}
R.~Sim and N.~Roy, ``Global a-optimal robot exploration in slam,'' in
  \emph{Proceedings of the 2005 IEEE international conference on robotics and
  automation}.\hskip 1em plus 0.5em minus 0.4em\relax IEEE, 2005, pp. 661--666.

\bibitem{placed2022general}
J.~A. Placed and J.~A. Castellanos, ``A general relationship between optimality
  criteria and connectivity indices for active graph-slam,'' \emph{IEEE
  Robotics and Automation Letters}, vol.~8, no.~2, pp. 816--823, 2022.

\bibitem{placed2022explorb}
J.~A. Placed, J.~J.~G. Rodr{\'\i}guez, J.~D. Tard{\'o}s, and J.~A. Castellanos,
  ``Explorb-slam: Active visual slam exploiting the pose-graph topology,'' in
  \emph{Iberian Robotics conference}.\hskip 1em plus 0.5em minus 0.4em\relax
  Springer, 2022, pp. 199--210.

\bibitem{barber1996quickhull}
C.~B. Barber, D.~P. Dobkin, and H.~Huhdanpaa, ``The quickhull algorithm for
  convex hulls,'' \emph{ACM Transactions on Mathematical Software (TOMS)},
  vol.~22, no.~4, pp. 469--483, 1996.

\bibitem{kummerle2011g2o}
R.~K{\"u}mmerle, G.~Grisetti, H.~Strasdat, K.~Konolige, and W.~Burgard, ``g2o:
  A general framework for graph optimization,'' in \emph{2011 IEEE
  International Conference on Robotics and Automation}.\hskip 1em plus 0.5em
  minus 0.4em\relax IEEE, 2011, pp. 3607--3613.

\bibitem{mladen2016nfr}
W.~B. Mladen~Mazuran and G.~D. Tipaldi, ``Nonlinear factor recovery for
  long-term slam,'' \emph{The International Journal of Robotics Research}, p.
  0278364915581629, 2015.

\bibitem{carlevaris2014generic}
N.~Carlevaris-Bianco, M.~Kaess, and R.~M. Eustice, ``Generic node removal for
  factor-graph slam,'' \emph{IEEE Transactions on Robotics}, vol.~30, no.~6,
  pp. 1371--1385, 2014.

\end{thebibliography}
\addtolength{\textheight}{-12cm}   







\end{document}